\documentclass[11pt,a4paper]{article}
\usepackage[hyperref]{emnlp2020}
\usepackage{times}
\usepackage{latexsym}

\usepackage{microtype}
\usepackage{amsmath}
\usepackage{amsfonts,amssymb}
\usepackage{multirow}
\usepackage{graphicx}
\usepackage{booktabs}
\usepackage{subcaption, caption}

\aclfinalcopy 
\def\aclpaperid{1328} 

\title{\textsc{StyleDGPT}: Stylized Response Generation with Pre-trained Language Models}

\author{
  Ze Yang$^1$, Wei Wu$^2$, Can Xu$^3$, Xinnian Liang$^1$, Jiaqi Bai$^1$, \\ \textbf{Liran Wang}$^1$, \textbf{Wei Wang}$^4$, and \textbf{Zhoujun Li}$^1$\thanks{~~~Corresponding Author}~~~~~\\
  $^1$State Key Lab of Software Development Environment, Beihang University, Beijing, China\\
  $^2$Meituan, Beijing, China \;\;
  $^3$Microsoft, Beijing, China \;\;
  $^4$China Resources Group\\
  \texttt{\{tobey,xnliang,bjq,wanglr,lizj\}@buaa.edu.cn}\\
  \texttt{\{wuwei19850318,ww.cs.tj\}@gmail.com} \;
  \texttt{caxu@microsoft.com}
}
\date{}

\begin{document}
\maketitle
\begin{abstract}
  Generating responses following a desired style has great potentials to extend applications of open-domain dialogue systems, yet is refrained by lacking of parallel data for training. In this work, we explore the challenging task with pre-trained language models that have brought breakthrough to various natural language tasks. To this end, we introduce a KL loss and a style classifier to the fine-tuning step in order to steer response generation towards the target style in both a word-level and a sentence-level. Comprehensive empirical studies with two public datasets indicate that our model can significantly outperform state-of-the-art methods in terms of both style consistency and contextual coherence.

\end{abstract}

\section{Introduction}
With advances in neural machine learning \cite{sutskever2014sequence,gehring2017convolutional,vaswani2017attention} and availability of huge amount of human conversations on social media, there has been significant progress on building open-domain dialogue systems with natural language generation techniques.  Though neural generative models are notorious for replying with bland responses \cite{li2015diversity}, some very recent work demonstrates that response generation models learned with pre-training techniques \cite{radford2019language} can effectively overcome the deficiency suffered by previous models and are capable of having smooth conversations with humans through reasonable and specific replies \cite{wolf2019transfertransfo,zhang2019dialogpt}.  

The compelling performance exhibited by the pre-trained dialogue models encourages us to explore more difficult yet important problems in conversational AI. In this work, we study stylized response generation, that is responses provided by a model should not only be coherent with the conversation contexts, but also be consistent with a designated style. Such research could facilitate developers to customize their dialogue systems in terms of response styles, and thus broaden applications of the systems, from a social companion \cite{shum2018eliza} or a virtual assistant \cite{ram2018conversational} to a variety of vertical scenarios such as customer service (requiring a polite style),  virtual characters in games (requiring specific personas), assistants in specific domains (requiring domain knowledge), etc. Normally, a target style is specified by a non-conversational corpus (e.g., novels, news, blogs, etc.) apart from the paired dialogue corpus \cite{luan-etal-2017-multi,niu-bansal-2018-polite,gao-etal-2019-structuring}. Thus, the major challenge of the task lies in the scarcity of paired data for learning the correspondence between conversation contexts and proper responses in the desired style, which is a key factor in success of the neural dialogue models developed so far. As a result, it is very likely that a response either digresses from the context of the current dialogue \cite{luan-etal-2017-multi,gao-etal-2019-structuring}, or loses fidelity to the target style \cite{niu-bansal-2018-polite}.

We consider addressing the challenge by taking advantage of the large scale pre-trained language models. The basic idea is that deep neural language models learned from huge amount of text, such as GPT-2 \cite{radford2019language} and DialoGPT \cite{zhang2019dialogpt}, have packed enough style knowledge into their parameters \cite{Dathathri2020Plug}, and thus by simply steering the distribution in decoding towards the desired style, we can obtain both contextual coherence and style consistency. Following the idea, we build a response generation model on top of a pre-trained language model and devise both a word-level loss and a sentence-level loss to fine-tune the pre-trained model towards the target style. The word-level loss regularizes the likelihood of response generation with a KL divergence term between the probability of dialogues and the probability of stylized language estimated by fine-tuning a pre-trained language model on the style corpus, while the sentence-level loss maximizes the likelihood of a response given by the pre-trained response generation model being classified as a sentence matching the target style.  We employ a Gumbel trick to overcome the obstacle in back-propagation due to the discrete nature of natural language when optimizing the sentence-level loss. The final response is selected by a sample-and-rank strategy to further enhance relevance regarding to the dialogue context and fidelity regarding to the target style. 

We name our model \textsc{StyleDGPT} standing for ``Stylized DialoGPT''.
Empirical studies are conducted on two tasks: arXiv-style response generation and Holmes-style response generation with the data shared in \cite{gao-etal-2019-structuring}, where responses in the style of scientific papers and the style of Sherlock Holmes novels are pursued respectively for a given context. Besides the style intensity used in \cite{gao-etal-2019-structuring}, we further examine style consistency from both a lexical perspective and a syntactic perspective with two new metrics. Evaluation results on both automatic metrics and human judgment indicate that our model can significantly outperform state-of-the-art methods. The code is available at \url{https://github.com/TobeyYang/StyleDGPT}.

Our contributions are three-fold: 
(1) proposal of tackling the problem of stylized response generation with pre-trained language models;
(2) proposal of a word-level objective and a sentence-level objective in fine-tuning of a pre-trained language model for the task;
and (3) empirical verification of the effectiveness of the proposed method on public datasets.
\section{Related Work}
\paragraph{Open-domain Dialogue Generation} has received more and more attention in NLP community.
Inspired by neural machine translation, early works apply the sequence-to-sequence model to this task and achieve promising results \cite{ritter2011data,shangL2015neural,vinyals2015neural}. 
Since then, various architectures have been proposed to address the key challenges in open-domain dialogue systems, including suppressing the generic responses \cite{li2015diversity,zhao2017learning,xing2017topic}, context modeling \cite{serban2016building,serban2017hierarchical,xing2017hierarchical,zhang2019recosa}, controlling the attributes of responses \cite{xu2019neural,zhou2017emotional,zhang2018learning,wang2018learning,see2019makes} and incorporating different types knowledge into generation \cite{li2016persona, zhang2018personalizing, zhou2017emotional, zhao2020low}. In this work, we study the problem of stylized response generation, which aims to incorporate the style information from non-parallel data into the generation process.

\paragraph{Stylized Text Generation} has attracted broad interest in recent years, especially the style transfer, which aims to alter one or more attributes of text while preserving the content.
A prevalent idea of unsupervised style transfer is learning to separate ``content'' and ``style'' of text and manipulate the style to induce transfer at inference time \cite{li-etal-2018-delete,AAAI1817015,john-etal-2019-disentangled}. 
However, some works show that the disentanglement cannot be met and is not necessary, and leverage techniques like reconstruction and back-translation introduced in unsupervised machine translation \cite{lample2018multiple}, transformer \cite{dai-etal-2019-style} to achieve unsupervised style transfer.
Different from style transfer, stylized response generation requires that the response is coherent with its context and the content can be varied. 
\citet{akama-etal-2017-generating} first train a basic model on a large-scale dialogue corpus and then fine-tune the model with a small stylized corpus. \citet{niu-bansal-2018-polite} propose three weakly-supervised methods to generate polite responses using non-parallel data. \citet{gao-etal-2019-structuring} build a structured latent space sharing between conversation modeling and style transfer. However, limited by the sparsity of the latent space, it is difficult to balance the style and contextual coherence while sampling in the neighborhood of the latent code of context at inference time.

\paragraph{Pretraining Methods} have led remarkable success in various NLP tasks which demonstrates its great capabilities in language understanding and text generation \cite{radford2018improving, radford2019language,devlin-etal-2019-bert,yang2019xlnet,liu2019roberta,conneau2019cross,clark2020electra}. 
Recently, the pretraining methods have also been used to tackle the key challenges in dialogue systems such as context representation \cite{mehri-etal-2019-pretraining}, response selection \cite{henderson2019convert}, knowledge-grounded response generation \cite{zhao2020low} and personalized response generation \cite{zheng2019pretraining}. In particular, the large-scale pre-trained open-domain dialogue systems \cite{zhang2019dialogpt,adiwardana2020towards} make a large step towards human-like chatbot against previous works which rely on complex frameworks developed over many years. On this basis, we propose to study the open-domain stylized response generation with pre-trained models in this work.

\section{Problem Formalization}
Suppose that we have a dialogue corpus $\mathcal{D}_{conv} = \{(X_i, Y_i)\}^n_{i=1}$ and a style corpus $\mathcal{D}_{style}=\{S_i\}^m_{i=1}$, where $\forall (X_i, Y_i) \in \mathcal{D}_{conv}$, $X_i$ is a conversation context and $Y_i$ a response to $X_i$, and $\forall S_i \in \mathcal{D}_{style}$, $S_i$ is a piece of text in the target style $\mathcal{S}$. We do not assume that there exists pairs $\{(X,Y^{\prime})\}$ with $Y^{\prime}$ expressed in the style $\mathcal{S}$\footnote{Some pairs in $\mathcal{D}_{conv}$ may meet the condition, but there is not an oracle that can tell us the information.}, and $\mathcal{D}_{style}$ could be collected from text in an arbitrary style (e.g. scientific papers, novels, etc.). Our goal is to learn a generation model $P(Y|X,\mathcal{S})$ with both $\mathcal{D}_{conv}$ and $\mathcal{D}_{style}$, and thus given a new context $X$, one can generate a response $Y$ that properly replies to the context $X$ following the style $\mathcal{S}$.

\section{Approach}

We employ DialoGPT \cite{zhang2019dialogpt} as the general response generation model $P(Y|X)$, and try to bias $P(Y|X)$ towards the language distribution $P(S)$ estimated from $\mathcal{D}_{style}$ in fine-tuning. Below, we first briefly review the OpenAI GPT-2 \cite{radford2019language} and DialoGPT, which serve as the backbone of our model. Then, we introduce two learning objectives from both a word perspective and a sentence perspective to interpolate style $\mathcal{S}$ into response generation.

\subsection{Backbone Networks}
GPT-2 is a large transformer based generative model pre-trained with language modeling \cite{radford2019language}. Given a sequence $X = (x_0, \cdots, x_n)$, the generative probability $p(X)$ can be factorized as the product of conditional probabilities over the tokens \cite{jelinek1980interpolated,bengio2003neural}:
\begin{equation}
    \label{eq:gpt-prob}
    p(X) = p(x_0)\prod^n_{i=1}p(x_i | x_0,\cdots, x_{i-1})
\end{equation}
GPT-2 uses a multi-layer transformer to model the distributions in a recurrent way. At step $t$, let us define $\mathbf{H}_t = [(\mathbf{K}_t^{(1)},\mathbf{V}_t^{(1)}), \cdots, (\mathbf{K}_t^{(l)}, \mathbf{V}_t^{(l)})]$ as the past key-value matrices where $(\mathbf{K}_t^{(i)}, \mathbf{V}_t^{(i)})$ represents the key-value pairs computed by the $i$-$th$ layer from step $0$ to step $t$, then given the input token $x_t$, the distribution of the next token $x_{t+1}$ can be efficiently calculated using the cached $\mathbf{H}_t$ which is formulated as:
\begin{equation}
    \label{eq:next_prob}
    \begin{split}
        &e_{x_t} = \mathbf{E} \, x_t^*,\\
        &o_{x_{t+1}}, \mathbf{H}_{t+1} = \mathrm{Transformer}(e_{x_t}, \mathbf{H}_t),\\
        &p(x_{t+1}| x_0,\cdots,x_{t}) = \mathrm{softmax}(\mathbf{W}_o \, o_{x_{t+1}}),
    \end{split}
\end{equation}
where $\mathbf{E} \in \mathbb{R}^{d_e\times |V|}$ is the word embedding matrix with $d_e$ the dimension and $|V|$ the vocabulary size, $x_t^*\in \mathbb{R}^{|V|}$ is a one-hot vector corresponding to token $x_t$, $o_{x_{t+1}} \in \mathbb{R}^{d_c}$ is the hidden state at step $t$ with $d_c$ the hidden size, and $\mathbf{W}_o \in \mathbb{R}^{|V|\times d_c}$ is a parameter matrix that maps the hidden state $o_{x_{t+1}}$ to a logit vector in the size of $|V|$. At inference time, $x_{t+1}$ is predicted following $p(x_{t+1}| x_0,\cdots,x_{t})$. Moreover, GPT-2 can also be used for language understanding. In this scenario, $o_X = (o_{x_1}, \cdots, o_{x_{n+1}})$ are treated as the representations of sequence $X$.

DialoGPT is a large conversational response generation model trained on $147$M conversation-like exchanges from Reddit community \cite{zhang2019dialogpt}. It inherits from GPT-2 and frames the response generation task as language modeling. For a context-response pair $(X, Y)$, a special token $\mathtt{\langle|endoftext|\rangle}$ is appended at the end of each dialogue turn and then all turns are concatenated into a long sequence. Let $M$ denote the length of the context sub-sequence and $(x_0,\cdots,x_{M-1},\cdots, x_N)$ denote the dialogue sequence after concatenation, the conditional generation probability of response $Y$ is defined as:
\begin{equation}
    \label{eq:dialogpt-prob}
    p(Y|X) = \prod^N_{i=M} p(x_i | x_0,\cdots, x_{i-1}).
\end{equation}

\subsection{Response Style Controlling}
\paragraph{Word-Level Objective} encourages the pre-trained response generation model $P(Y|X)$ (i.e. DialoGPT) to pick words expressing the desired style $\mathcal{S}$ in decoding. Specifically, we train a language model $P(S)$ with $\mathcal{D}_{style}$ on the basis of GPT-2 and use it as regularization to drive $P(Y|X)$ towards $P(S)$. 
It is inspired that if a response $Y$ is not consistent with the style $\mathcal{S}$, it will get high perplexity (i.e. $Y$ is far from the language space of $\mathcal{S}$).
Furthermore, $P(S)$ could not only provide an overall evaluation on the fidelity of a response $Y$, but also assign a direct probability distribution over the vocabulary at each step and thus provide word-level information about which words need to be promoted in generation.

For each  $(X, Y) \in \mathcal{D}_{conv}$, we denote $p_Y = (p_{y_1}, \cdots, p_{y_m})$ ($m$ is the length of $Y$) as the next-word distributions of $Y$ given by  $P(Y|X)$. Meanwhile, we feed $Y$ into $P(S)$ and obtain the next-word distributions $\hat{p}_Y = (\hat{p}_{y_1}, \cdots, \hat{p}_{y_m})$. Then the word-level objective is formulated as:
\begin{equation}
    \label{eq:word-loss}
    \mathcal{L}_{w} = \mathbb{E}_{(X,Y)\sim \mathcal{D}_{conv}} \, d(p_Y  \| \, \hat{p}_Y),
\end{equation}
where $d(p_Y  \| \, \hat{p}_Y)$ could be any metrics measuring the \textit{distance} between $p_Y$ and $\hat{p}_Y$. Here, we specify $d(\cdot \| \, \cdot)$ as the Kullback-Leibler (KL) divergence. Then,  $d(p_Y  \| \, \hat{p}_Y) = \sum^m_{i=1} \text{D}_{KL}(p_{y_i}  \| \, \hat{p}_{y_i})$.
At each step, $\mathcal{L}_w$ modifies the next-word distribution in the direction of $P(\mathcal{S})$ where the probabilities of words with the desired style $\mathcal{S}$ will be increased, which can encourage the selection of these words at inference time.

\paragraph{Sentence-level Objective} modifies $P(Y|X)$ towards the target style $\mathcal{S}$ from a syntactic and semantic perspective. In training, we hope that a response matching style $\mathcal{S}$ could have more impact in guiding the optimization of $P(Y|X)$ towards the desired direction. To this end, we first train a discriminative model $P(\mathcal{S}|X)$ to predict whether the input sequence $X$ matches the style $\mathcal{S}$. Formally, given an input sequence $X = (x_0, \cdots, x_n)$, the probability is defined as:
\begin{equation}
    \label{eq:discriminator}
    \begin{split}
        &p(\mathcal{S}|X) = \mathrm{sigmoid} (\mathbf{W}_{d} \ \hat{o}_X),\\
        &\hat{o}_X = \mathrm{average\_pooling}(o_X),
    \end{split}
\end{equation}
where $o_X=(o_{x_1}, \cdots, o_{x_{n+1}})$ are the representations of $X$ encoded by GPT-2, $\mathrm{average\_pooling}(\cdot)$ denotes the average pooling layer where the $i$-th element ${\hat{o}_X}^{(i)}$ is given by  $\frac{1}{n+1}\sum^{n+1}_{j=1} o_{x_j}^{(i)}, i\in [1,d_c]$, and $\mathbf{W}_d \in \mathbb{R}^{1\times d_c}$ is a parameter. In the training phase, positive examples are sampled from $\mathcal{D}_{style}$ while negative examples are utterances sampled from $\mathcal{D}_{conv}$ \footnote{The ratio of the positive and the negative is $1:5$ in our experiments.}. Then the sentence-level objective is formulated as:
\begin{equation}
    \label{eq:sentence-loss}
    \mathcal{L}_{s} = \mathbb{E}_{{(X,Y)\sim \mathcal{D}_{conv}},\,\widetilde{Y}\sim P(\widetilde{Y}|X)}[-\text{log} \, p(\mathcal{S}|\widetilde{Y})].
\end{equation}
$\mathcal{L}_{s}$ aims to regularize the output of the generation model by ascending the probability given by the discriminative model $P(\mathcal{S}|X)$, which is similar to the optimization process of the generator in GANs \cite{goodfellow2014generative}. The challenge is that since $\widetilde{Y}$ is discrete, it is impossible to back-propagate through sampling from $P(\widetilde{Y}|X)$. Although it can be circumvented by using the reinforcement learning (RL) algorithm \cite{sutton2000policy}, the performance is not satisfactory in our experiments. In this work, we propose using the Gumbel trick \cite{jang2016categorical} to tackle the challenge. At step $t$, instead of sampling a token from $p(x_{t+1}|x_0,\cdots,x_t)$, the input vector of step $t+1$ is obtained by:
\begin{equation}
    \label{eq:gumbel}
    x_{t+1}^* = \mathrm{gumbel\_softmax}(\mathbf{W_o}\ o_t, \tau),
\end{equation}
where $\tau$ is the temperature and when $\tau \rightarrow 0$, $x_{t+1}^* \in \mathbb{R}^{|V|}$ becomes a one-hot vector.

\paragraph{Training Objective.} The two objectives presented above are able to drive $P(Y|X)$ to generate responses with desirable style $\mathcal{S}$, but it will quickly result in irrelevant responses as both of them only focus on responses. To overcome this, we preserve the negative log-likelihood (NLL) loss in DialoGPT to maintain the relevance between the context and response:
\begin{equation}
    \label{eq:nll-loss}
    \mathcal{L}_{NLL}   = \mathbb{E}_{(X,Y) \sim \mathcal{D}_{conv}}[-\text{log} \, p(Y|X)]
\end{equation}

The final training loss is the weighted sum of the word-level loss, sentence-level loss, and relevance loss:
\begin{equation}
    \label{eq:final-loss}
    \mathcal{L} = \lambda_w \cdot \mathcal{L}_w + \lambda_s \cdot \mathcal{L}_s + \lambda_{NLL} \cdot \mathcal{L}_{NLL},
\end{equation}
where $\lambda_w$, $\lambda_s$, $\lambda_{NLL}$ are three weight scalars.

\paragraph{Sampling and Ranking.}
Because it is possible to generate non-stylized responses at inference time, we employ the sample-and-rank decoding strategy following \citet{gao-etal-2019-structuring}. First, we sample $N$ independent candidate responses for each context by using top-$k$ sampling method with temperature $T$. Then, we re-rank them in terms of both relevance and style intensity and select the candidate with the highest score as the final response. The score of a candidate $Y_i$ for context $X$ is defined as
\begin{equation}
    \label{eq:rank}
    score(Y_i) = \beta \cdot p(Y_i|X) + (1-\beta) \cdot p(\mathcal{S}|Y_i),
\end{equation}
where $p(Y_i|X)$ measures relevance of $Y_i$ regarding to $X$, $p(\mathcal{S}|Y_i)$ returns style intensity of $Y_i$ defined by the discriminative model $P(\mathcal{S}|X)$, and $\beta$ is a hyper-parameter.
\section{Experiments}
\subsection{Datasets}
In order to verify the effectiveness of our model, we experiment on two tasks: generating arXiv-style and Holmes-style responses. The statistics of datasets are summarized in Table \ref{table:datasets}.
The datasets are constructed following the pipeline in \citet{gao-etal-2019-structuring}. 
The style corpus $\mathcal{D}_{style}$ for arXiv-style response generation task consists of {\raise.17ex\hbox{$\scriptstyle\mathtt{\sim}$}}$1$M sentences that are extracted from the LaTex source code of papers on website arXiv.org from 1998 to 2002 \footnote{downloaded from \url{http://www.cs.cornell.edu/projects/kddcup/datasets.html}}. 
For Holmes-style response generation task, $\mathcal{D}_{style}$ contains {\raise.17ex\hbox{$\scriptstyle\mathtt{\sim}$}}$38$k sentences built from ebooks of Sherlock Holmes novel series downloaded from the site Gutenberg.org \footnote{\url{http://www.gutenberg.org}}. 
Both tasks share the same conversation dataset $\mathcal{D}_{conv}$ which consists of $10$M context-response pairs extracted from user posts and comments on site Reddit.com during the year $2011$ \footnote{We use the raw data collected by a third party \url{http://files.pushshift.io/reddit}.}. 
The validation set $\mathcal{D}_{val}$ and the test set $\mathcal{D}_{test}$ are constructed by filtering the Reddit data in $2013$ with the classifier in \cite{gao-etal-2019-structuring}   (intensity score $> 0.4$) \footnote{available at \url{https://github.com/golsun/StyleFusion/tree/master/classifier}}. 
As \citet{gao-etal-2019-structuring} do not release their test data, nor specify the size of the test set, we randomly select 2k/2k samples as the validation/test sets, and each context has at least $4$ responses.
\begin{table}[ht!]
    \centering
    \small
    \resizebox{\linewidth}{!}{
    \begin{tabular}{c|cc|cc}
    \hline
    \multirow{2}{*}{Task} & \multicolumn{2}{c|}{Training} & Validation & Test \\
          & $\mathcal{D}_{conv}$ & $\mathcal{D}_{style}$ & $\mathcal{D}_{val}$ & $\mathcal{D}_{test}$\\
    \hline
    \multirow{2}{*}{arXiv-style} & Reddit & arXiv & \multicolumn{2}{c}{arXiv-style Reddit} \\
                & 10,000,000 & 1,347,538 & 2,000 & 2,000\\
    \hline
    \multirow{2}{*}{Holmes-style} & Reddit & Holmes & \multicolumn{2}{c}{Holmes-style Reddit} \\
                & 10,000,000 & 38,309 & 2,000 & 2,000\\
    \hline
    \end{tabular}
    }
    \caption{Tasks and datasets}
    \label{table:datasets}
\end{table}

\subsection{Evaluation Methodology}
We compare different models with both automatic metrics and human judgment.
\paragraph{Automatic Metrics.} For automatic evaluation, we measure the quality of generated responses from three aspects: \textbf{Style Consistency}, \textbf{Relevance}, and \textbf{Diversity}. The relevance is measured with BLEU \cite{papineni2002bleu} and Rouge \cite{lin2004rouge} \footnote{Both metrics are computed by scripts of a public NLG evaluation project available at \url{https://github.com/Maluuba/nlg-eval}.}. To evaluate diversity, we follow \citet{li2015diversity} and use Distinct-1 (Dist-1) and Distinct-2 (Dist-2) as metrics which are calculated as ratios of distinct unigrams and bigrams in responses, respectively. 

In terms of style consistency, existing work only measures the style intensity using classifiers \cite{gao-etal-2019-structuring}. 
However, the style of text is an amalgam, and differences between two styles are reflected in multiple linguistic dimensions \cite{verma2019lexical}. 
Thus, we propose to evaluate the style of response from three perspectives:
(1) \textbf{\textit{Intensity}}: we report the scores from the discriminative model $p(\mathcal{S}|X)$\footnote{The evaluation is more accurate than that from the classifiers available at \url{https://github.com/golsun/StyleFusion/tree/master/classifier}
because of the capability of GPT-2.}.
(2) \textbf{\textit{Lexical}}:  it is a word-level metric that measures the distance between two lexical distributions. We first build a lexicon with all the ngrams ($N=1,2,3,4$) from $\mathcal{D}_{conv}$ and $\mathcal{D}_{style}$ (i.e., Reddit, arXiv, and Holmes corpora). To reduce noise, ngrams that occur less than $10$ times are filtered out and there are $1,346,175$ distinct ngrams left. Then the lexical distributions of a model and the target style can be represented as normalized $1,346,175$-dimensional vectors with each element the frequency of the corresponding ngram in the generated responses (over the test set) and $\mathcal{D}_{style}$ respectively. Finally, we calculate the Jensen-Shannon divergence \cite{jsd2004} to measure the distance of the two vectors.
(3) \textbf{\textit{Syntactic}}: it is a sentence-level metric. Motivated by \citet{feng-etal-2012-characterizing}, the style of text can be recognized by the ratio of the following $5$ syntactic types: (a) \textit{simple}; (b) \textit{compound}; (c) \textit{complex}; (d) \textit{complex-compound}; (e) \textit{others}. The type of a sentence is determined by the algorithm proposed by \citet{feng-etal-2012-characterizing} which relies on the PCFG tree parsed by the Stanford CoreNLP \footnote{\url{https://stanfordnlp.github.io/CoreNLP}}. We compute the distributions of the style corpus and responses generated by models and report the Jensen-Shannon divergence.

\begin{table*}[tp!]
  \centering
  \small
  \resizebox{\linewidth}{!}{
    \begin{tabular}{l|ccc|ccc|cc}
  \hline
  \hline
  \multicolumn{1}{c|}{\multirow{2}[2]{*}{Models}} & \multicolumn{3}{c|}{Style Consistency} & \multicolumn{3}{c|}{Relevance ($\uparrow$)} & \multicolumn{2}{c}{Diversity ($\uparrow$)} \\
          & Intensity ($\uparrow$) & Lexical ($\downarrow$) & Syntactic ($\downarrow$) & BLEU1 & BLEU2 & RougeL & Dist-1 & Dist-2 \\
  \hline
  \multicolumn{9}{c}{arXiv-style Response Generation}\\
  \hline
    MTask \cite{luan-etal-2017-multi} & 0.284 & 0.7565 & 0.2653 & 13.42 & 3.56  & 11.53 & 0.040 & 0.091 \\
    S2S+LM \cite{niu-bansal-2018-polite} & 0.399 & 0.7484 & 0.2549 & 15.25 & 4.62 & 10.41 & 0.052 & 0.273 \\
    StyleFusion \cite{gao-etal-2019-structuring} & 0.412 & 0.7582 & 0.2282 & 16.81 & 5.69  & 10.82 & 0.055 & 0.107 \\
    DialoGPT \cite{zhang2019dialogpt}  & 0.208 & 0.6518 & 0.2561 & 17.84 & 5.20  & 10.68 & \textbf{0.296} & \textbf{0.711} \\
    \textsc{StyleDGPT}   & \textbf{0.503} & \textbf{0.6237} & \textbf{0.1912} & \textbf{19.04} & \textbf{5.74}  & \textbf{12.49} & 0.228 & 0.614 \\
  \hline
  \multicolumn{9}{c}{Holmes-style Response Generation}\\
  \hline
    MTask \cite{luan-etal-2017-multi}  & 0.276 & 0.7106 & 0.2356 & 24.47 & 8.87  & 16.03 & 0.027 & 0.063 \\
    S2S+LM \cite{niu-bansal-2018-polite}  & 0.450 & 0.5982 & 0.1959 & 25.32 & 9.15 & 14.82 & 0.051 & 0.304 \\
    StyleFusion \cite{gao-etal-2019-structuring} & 0.479 & 0.7023 & 0.1946 & 25.91 & 9.68  & 15.87 & 0.045 & 0.098 \\
    DialoGPT \cite{zhang2019dialogpt}  & 0.282 & 0.5814 & 0.1598 & 27.19 & 8.31  & 14.78 & \textbf{0.172} & \textbf{0.589} \\
    \textsc{StyleDGPT} & \textbf{0.602} & \textbf{0.4807} & \textbf{0.0861} & \textbf{29.58} & \textbf{10.15} & \textbf{17.10} & 0.101 & 0.452 \\
  \hline
  \hline
    \end{tabular}%
  }
  \caption{Evaluation results on automatic metrics. Numbers in \textbf{bold} indicate the best performing models under the corresponding metrics. $\uparrow$/$\downarrow$ means higher/lower values are better, respectively. The unit for relevance is percentage.}
  \label{tab:automatic_evaluation}%
\end{table*}%

\paragraph{Human Evaluation.} We recruit $3$ well-educated native speakers as annotators to compare our model with each of the baselines. Each annotator checks one context with two responses at a time with one response from our model and the other from a baseline model, and the two responses are shown in random order. The annotators then are asked to compare them on four aspects: (1) \textbf{Style Consistency}: if the response exhibits the desired style $\mathcal{S}$;
(2) \textbf{Fluency}: if the response is fluent without any grammatical errors; (3) \textbf{Relevance}: if the response is coherent with the given context; and (4) \textbf{Informativeness}: if the response is rich in content and thus could keep the conversation going. For each aspect, if the annotator cannot tell which response is better, he/she is asked to label a ``Tie''. For each task, $200$ test examples are sampled for annotation. Each pair of responses receive $3$ labels on each of the three aspects, and the agreement among the annotators are measured by Fleiss' kappa \cite{fleiss1973equivalence}.

\subsection{Baselines}
We compare our model with the following baselines: 
(1) \textbf{MTask}: a vanilla multi-task learning model proposed by \citet{luan-etal-2017-multi} trained with both $\mathcal{D}_{conv}$ and $\mathcal{D}_{style}$. We use the code implemented by \citet{gao-etal-2019-structuring} included in the project \url{https://github.com/golsun/StyleFusion}. 
(2) \textbf{S2S+LM}: the fusion model proposed by \citet{niu-bansal-2018-polite} that merges the decoder of a seq2seq model trained on $\mathcal{D}_{conv}$ and a language model trained on $\mathcal{D}_{style}$ by weighted averaging the word distributions at inference time. We use the code published at \url{https://github.com/WolfNiu/polite-dialogue-generation}. 
(3) \textbf{StyleFusion}: the regularized multi-task learning model proposed by \citet{gao-etal-2019-structuring} which builds a structured latent space to bridge the conversation modeling and style transfer. The model is jointly learned with $\mathcal{D}_{conv}$ and $\mathcal{D}_{style}$. We run the code released at \url{https://github.com/golsun/StyleFusion} with default settings. 
(4) \textbf{DialoGPT}: an open-domain pre-trained response generation model built upon GPT-2 that attains a performance close to human \cite{zhang2019dialogpt}. We use the $345$M fine-tuned model which can be downloaded from \url{https://github.com/microsoft/DialoGPT}.

\begin{table*}[th!]
  \renewcommand{\arraystretch}{1.1}
      \small
      \centering
      \resizebox{0.98\textwidth}{!}{
      \begin{tabular}{l|ccc|ccc|ccc|ccc|c}
          \hline
          \hline
          \multicolumn{1}{c|}{\multirow{2}[2]{*}{Models}} & \multicolumn{3}{c|}{Style Consistency} &\multicolumn{3}{c|}{Fluency} & \multicolumn{3}{c|}{Relevance} & \multicolumn{3}{c|}{Informativeness} & \multicolumn{1}{c}{\multirow{2}[2]{*}{Kappa}}\\
          & W($\%$) & L($\%$) & T($\%$) & W($\%$) & L($\%$) & T($\%$) & W($\%$) & L($\%$) & T($\%$) & W($\%$) & L($\%$) & T($\%$)\\
          \hline
          \multicolumn{14}{c}{arXiv-style Response Generation}\\
          \hline
          \textsc{StyleDGPT} vs. MTask      & 43.6 & 25.2 & 31.2 & 25.5 & 20.0 & 54.5 & 31.3 & 20.5 & 48.2 & 37.4 & 20.0 & 43.6 & 0.62\\
          \textsc{StyleDGPT} vs. S2S+LM     & 41.7 & 21.6 & 36.7 & 39.0 & 7.8  & 53.2 & 53.3 & 10.3 & 36.4 & 38.2 & 17.3 & 44.5 & 0.67\\
          \textsc{StyleDGPT} vs. StyleFusion& 38.2 & 18.4 & 43.4 & 23.6 & 18.3 & 58.1 & 38.0 & 16.2 & 45.8 & 31.8 & 15.2 & 53.0 & 0.65\\
          \textsc{StyleDGPT} vs. DialoGPT   & 51.3 & 10.2 & 38.5 & 16.2 & 21.8 & 62.0 & 21.2 & 26.5 & 52.3 & 23.2 & 23.8 & 53.0 & 0.61\\
          \hline
          \multicolumn{14}{c}{Holmes-style Response Generation}\\
          \hline
          \textsc{StyleDGPT} vs. MTask      & 46.3 & 13.8 & 39.1 & 28.0 & 14.8 & 57.2 & 43.8 & 15.4 & 40.8 & 36.8 & 12.0 & 51.2 & 0.65\\
          \textsc{StyleDGPT} vs. S2S+LM     & 45.0 & 19.5 & 35.5 & 36.3 & 4.8  & 58.9 & 52.2 & 9.0  & 38.8 & 38.6 & 16.3 & 45.1 & 0.61\\
          \textsc{StyleDGPT} vs. StyleFusion& 36.2 & 18.0 & 45.8 & 31.4 & 11.5 & 57.1 & 36.0 & 17.5 & 46.5 & 41.3 & 12.2 & 46.5 & 0.70\\
          \textsc{StyleDGPT} vs. DialoGPT   & 52.0 & 13.3 & 34.7 & 14.4 & 12.6 & 73.0 & 19.3 & 20.5 & 60.2 & 22.6 & 15.8 & 61.6 & 0.63\\
          \hline
          \hline
      \end{tabular}
      }
      \caption{Human annotation results. W, L, and T refer to Win, Lose, and Tie, respectively. The ratios are calculated by combining labels from the three annotators.}
      \label{tab:manual_result}
\end{table*}

\begin{table*}[thp!]
  \centering
  \small
  \resizebox{0.98\linewidth}{!}{
    \begin{tabular}{l|ccc|ccc|cc}
  \hline
  \hline
  \multicolumn{1}{c|}{\multirow{2}[2]{*}{Models}} & \multicolumn{3}{c|}{Style Consistency} & \multicolumn{3}{c|}{Relevance ($\uparrow$)} & \multicolumn{2}{c}{Diversity ($\uparrow$)} \\
          & Intensity ($\uparrow$) & Lexical ($\downarrow$) & Syntactic ($\downarrow$) & BLEU1 & BLEU2 & RougeL & Dist-1 & Dist-2 \\
  \hline
  \multicolumn{9}{c}{arXiv-style Response Generation}\\
  \hline
    \textsc{StyleDGPT}   & 0.503 & 0.6237 & 0.1912 & 19.04 & 5.74  & 12.49 & 0.228 & 0.614 \\
     \textsc{StyleDGPT} (w/o $\mathcal{L}_w$)  & 0.378 & 0.6357 & 0.2165 & 18.66 & 5.69  & 11.84 & 0.260 & 0.651 \\
    \textsc{StyleDGPT} (w/o $\mathcal{L}_s$)  & 0.670 & 0.6213 & 0.2177 & 17.28 & 4.85  & 11.39 & 0.182 & 0.564 \\
    \textsc{StyleDGPT} (w/o $\mathcal{L}_{NLL}$) &0.880& 0.5712 & 0.1594 & 13.16 & 4.08  & 11.86 & 0.046 & 0.273\\
  \hline
  \multicolumn{9}{c}{Holmes-style Response Generation}\\
  \hline
    \textsc{StyleDGPT} & 0.602 & 0.4807 & 0.0861 & 29.58 & 10.15 & 17.10 & 0.101 & 0.452 \\
    \textsc{StyleDGPT} (w/o $\mathcal{L}_w$)   & 0.497 & 0.5007 & 0.1194 & 29.21 & 9.34  & 16.14 & 0.130 & 0.514 \\
    \textsc{StyleDGPT} (w/o $\mathcal{L}_s$)   & 0.680 & 0.4716 & 0.1551 & 27.89 & 9.22  & 16.54 & 0.097 & 0.459 \\
    \textsc{StyleDGPT} (w/o $\mathcal{L}_{NLL}$)  & 0.891 & 0.4709 & 0.1521 & 26.54 & 8.56  & 15.53 & 0.049 & 0.298 \\
  \hline
  \hline
    \end{tabular}%
  }
  \caption{Ablation results on automatic metrics.}
  \label{tab:ablation_automatic}%
\end{table*}%

\subsection{Implementation Details}
Our models are implemented with the Huggingface transformers repository \footnote{\url{https://github.com/huggingface/transformers}}. 
To balance cost and effect, the language model $P(\mathcal{S})$ and the discriminative model $P(\mathcal{S}|X)$ are built upon  GPT-2 (117M) with $12$ layers and $768$ hidden units. 
The embedding layer and the transformer module are shared between two models, and we only optimize the parameters of the projection layer and the classification layer, respectively. 
We choose DialoGPT (345M) as the basis of \textsc{StyleDGPT} which has $24$ layers and $1024$ hidden units.
In both tasks, we use the vocabulary published along with GPT-2 by OpenAI that contains $50,257$ tokens. 
The temperature $\tau$ of gumabel softmax is set as $0.1$. 
Hyper-parameters are selected via grid search, and $\lambda_w$/$\lambda_s$/$\lambda_r$ are finally set as $0.0005$/$0.05$/$1$ for the arXiv-style response generation task and $0.005$/$0.05$/$1$ for the Holmes-style response generation task, respectively. 
All models are trained with the Adam optimizer \cite{kingma2014adam} ($\beta_1=0.9$, $\beta_2=0.999$) with a learning rate of $5\times 10^{-7}$.
We choose $k=40$ and $T=1.0$ in top-$k$ decoding following \cite{radford2019language,adiwardana2020towards}.
At inference time, all approaches including our model and baselines generate $50$ candidates for each context (i.e. $N=50$), and the top one candidate is selected for evaluation according to Equation (\ref{eq:rank}). 

\subsection{Evaluation Results}
\paragraph{Automatic Evaluation.} Table \ref{tab:automatic_evaluation} reports the evaluation results on automatic metrics. Without any complicated manipulation on latent spaces, \textsc{StyleDGPT}  outperforms the non-pre-trained baselines with large margins on all metrics in both tasks, demonstrating the advantage of pre-training over the state-of-the-art method in stylized response generation. 
The significant improvement over the vanilla DialoGPT on style consistency indicates that \textsc{StyleDGPT} can effectively leverage the extra objectives and bias response decoding towards the desired style. Moreover, it seems that forcing responses to a particular style (i.e., arXiv style and Holmes style) is also helpful in relevance, though there is a sacrifice on diversity. This is because the search space in decoding now becomes more concentrated on words that can express the target styles\footnote{Note that human responses for calculating the relevance metrics are biased to the target styles according to a style classifier.}. 

\paragraph{Human Evaluation.}
Table \ref{tab:manual_result} reports the results of human evaluation. The values of kappa are all above $0.6$, indicating substantial agreement among the three annotators. We can see \textsc{StyleDGPT} outperforms all non-pre-trained baselines on the three aspects, which echoes the results of automatic evaluation.  
Specifically, S2S+LM achieves poor performance on fluency because the weighted average of the token distributions predicted by the language model and the seq2seq decoder harms their attributes of language modeling, which also leads to low relevance.
Compared to DialoGPT, we notice that \textsc{StyleDGPT} significantly improves upon style consistency while achieves comparable performance on relevance and informativeness, which demonstrates the effectiveness of the proposed objectives in fine-tuning. 

\subsection{Discussions}

\paragraph{Ablation Study.} 
To understand the roles of $\mathcal{L}_w$, $\mathcal{L}_s$, and $\mathcal{L}_{NLL}$ in learning to generate stylized responses, we remove them one at a time from the full objective in Equation (\ref{eq:final-loss}), and then check the performance of the variants of \textsc{StyleDGPT} on the test sets. Table \ref{tab:ablation_automatic} reports the evaluation results. We can see that (1) all the three objectives are useful, as removing any of them will cause a performance drop on some metrics; (2) $\mathcal{L}_w$ is more important to lexical consistency while $\mathcal{L}_s$ is more important to syntactic consistency, which echoes our motivation in design of the two objectives; and (3) without $\mathcal{L}_{NLL}$, the model will be misled by the style corpus and lose the connection with conversation contexts. 

\begin{figure}[t!]
  \includegraphics[width=\columnwidth]{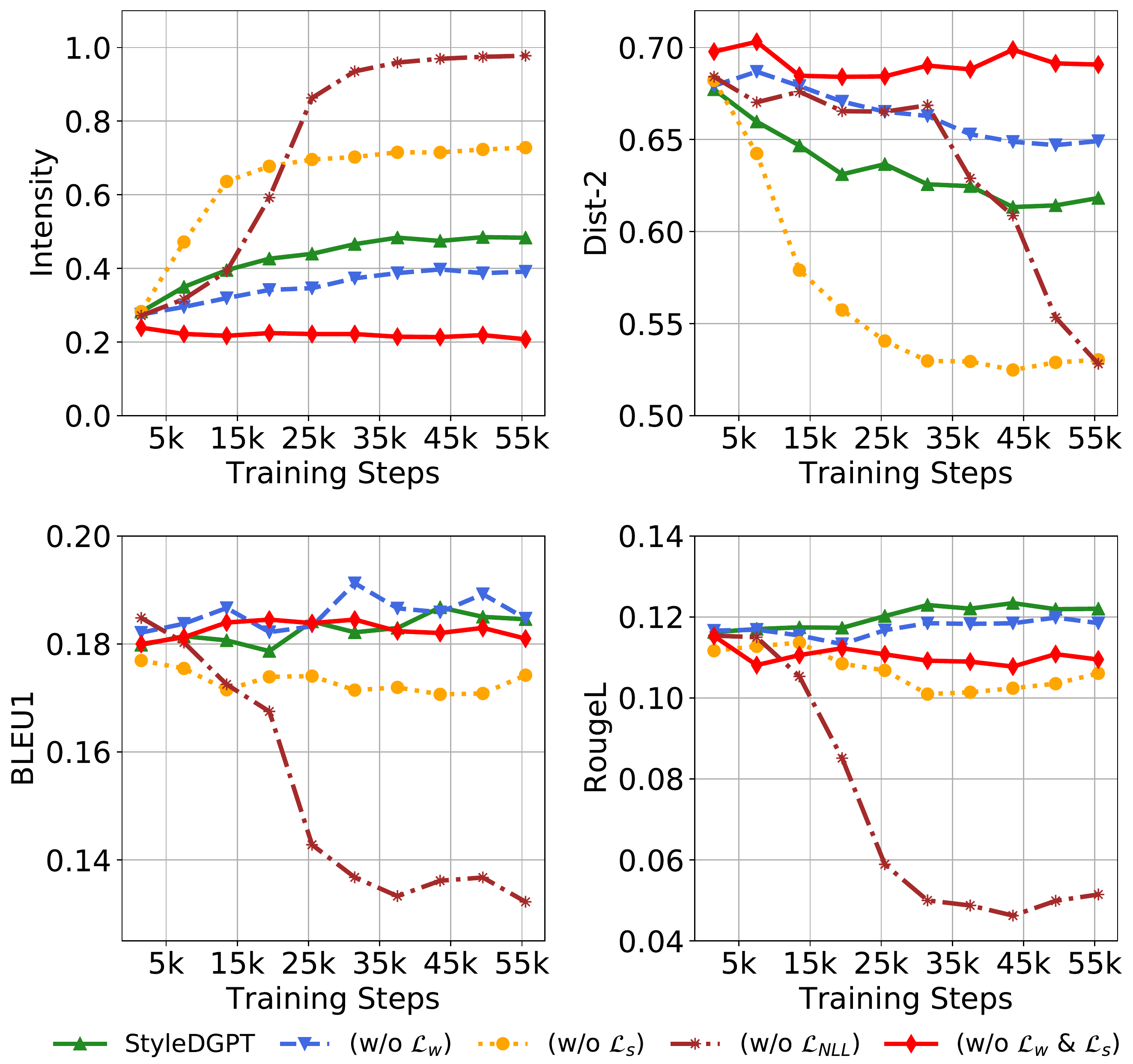}
  \caption{Trajectories of ablated \textsc{StyleDGPT} on the validation set of arXiv-style response generation.}
  \label{fig:analysis}
\end{figure}

Since $\mathcal{L}_w$, $\mathcal{L}_s$, and $\mathcal{L}_{NLL}$ are coordinated in learning of \textsc{StyleDGPT}, more insights about the effect of the objectives can be obtained by checking the trajectories of the variants on validation, as illustrated by Figure \ref{fig:analysis}\footnote{Similar trends are observed on Holmes-style response generation.}.  Without $\mathcal{L}_s$, there is a steady and significant improvement on style intensity but dramatic drops on BLEU1, RougeL, and Dist-2 (compared with the model without both $\mathcal{L}_s$ and $\mathcal{L}_w$), which indicates that $\mathcal{L}_w$ can provide stronger guidance regarding style expression than $\mathcal{L}_s$. On the other hand, comparing \textsc{StyleDGPT} w/o $\mathcal{L}_w$ and \textsc{StyleDGPT} w/o $\mathcal{L}_w$ \&  $\mathcal{L}_s$, we find that $\mathcal{L}_s$ can gradually and moderately improve upon style intensity and relevance with only a little hurt on diversity. Finally, when $\mathcal{L}_{NLL}$ is removed, the model will quickly forget conversation contexts and converge to the style language model.  The full model balances the effect of the three losses and attains both style consistency and contextual coherence, though it has to suffer from diversity drop due to the existence of $\mathcal{L}_w$.

\begin{figure}[t!]
  \begin{subfigure}[c]{\columnwidth}\label{fig:intensity}
    \includegraphics[width=1\textwidth]{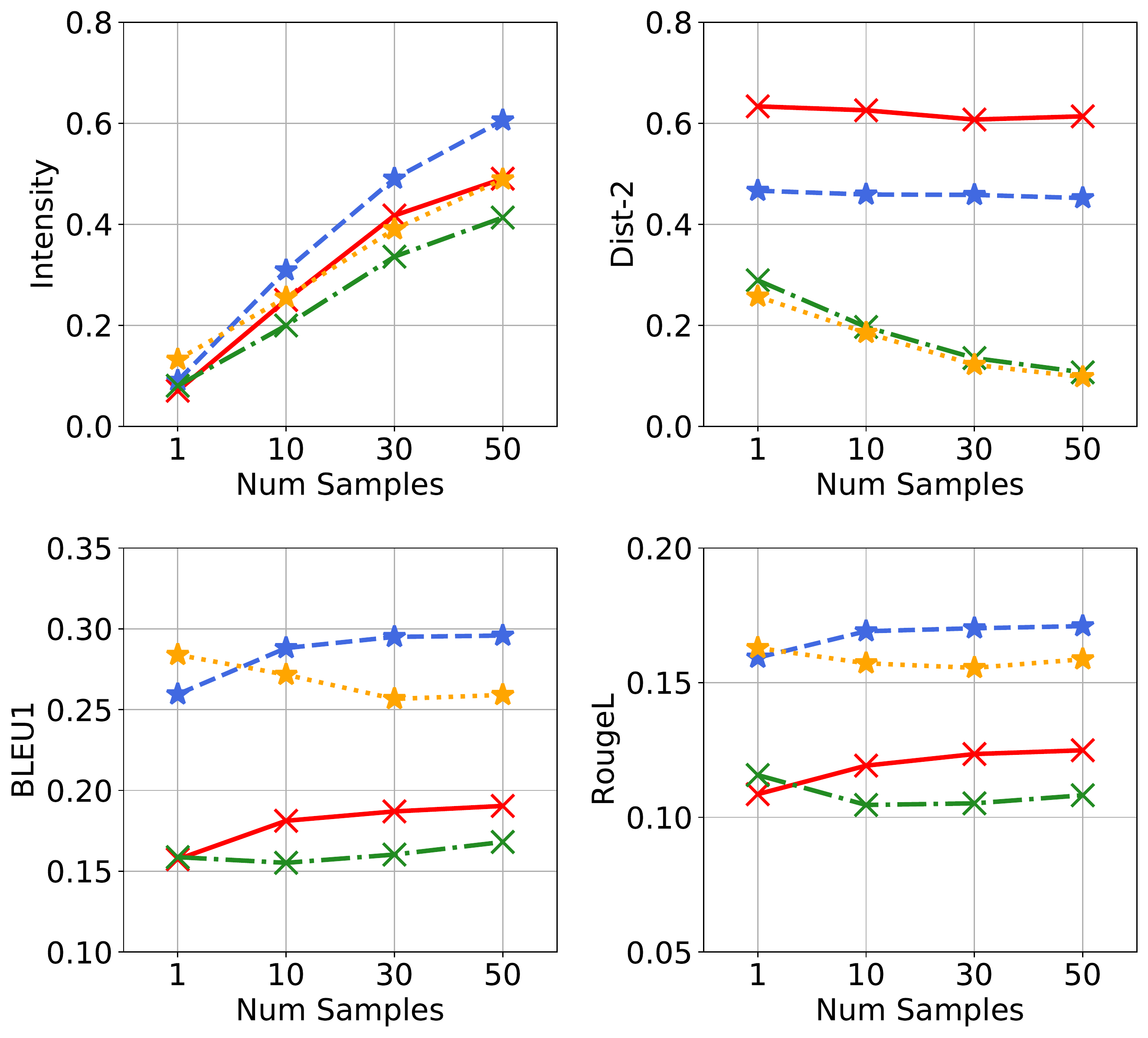}
  \end{subfigure}
  \begin{subfigure}[c]{\columnwidth}\label{fig:distinct}
  \centering
    \hspace{4mm}\includegraphics[width=0.6\textwidth]{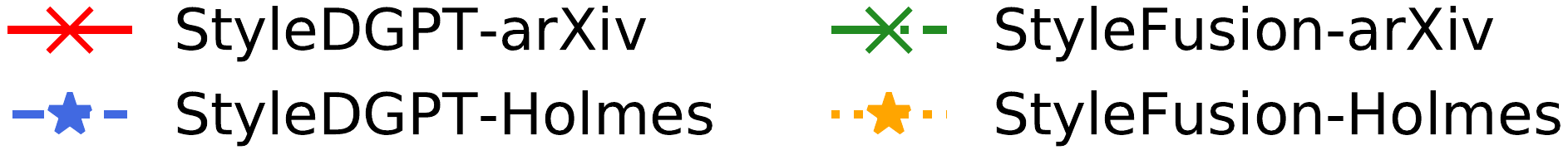}
  \end{subfigure}
  \caption{Comparisons over the number of sampled candidates on both tasks.}
  \label{fig:sampling_analysis}
\end{figure}

\paragraph{Impact of the Sampling Number $N$.}
To understand how the sample-and-rank strategy affects model performance, we evaluate \textsc{StyleDGPT} and StyleFusion by varying the sampling number $N$ in $\{1,10,30,50\}$ on both tasks. Figure \ref{fig:sampling_analysis} shows the results. We observe that (1) style intensity is more sensitive to the value of $N$ than other metrics; (2) though the two models are comparable in terms of style intensity when $N=1$, \textsc{StyleDGPT} can exhibit the desired styles with fewer samples; (3) \textsc{StyleDGPT} is always better than StyleFusion on Dist-2, thanks to DialoGPT; and (4) while \textsc{StyleDGPT} is able to attain both style consistency and contextual coherence with enough samples, it is difficult for StyleFusion to balance the two aspects, as when $N$ increases, both BLEU1 and RougeL drop. This is because when sampling in the neighborhood of the representation of a context in the structured latent space, reaching a stylized but less relevant point becomes easier when the number of samples increases.

\paragraph{Case Study.} Finally, we conduct qualitative analysis with some examples given in Table \ref{tab:example_arXiv} and Table \ref{tab:example_Holmes}. First, we find that the non-pre-trained models can generate interesting responses occasionally (e.g., ``the ring of fire'' and ``the first harry potter movie.'' in Table \ref{tab:example_Holmes}), but they are not always coherent with the contexts nor specific. 
In contrast, responses from the pre-trained models (i.e., DialoGPT and \textsc{StyleDGPT}) are highly coherent with the contexts and rich in content. 
In terms of style consistency, responses given by \textsc{StyleDGPT} are more stylized that contain more words specific to a certain style. For example, ``constant state'' and ``quantum fluctuations'' are concepts in physical science, ``the behavior'' of ``species'' is the research object of ethology, and ``accident'', ``investigate'', ``discover'' and ``trap'' are topical words in the Sherlock Holmes novel series.

\begin{table}[ht]
  \centering
  \small
  \begin{tabular}{|p{0.08\textwidth}|p{0.34\textwidth}|} 
    \hline 
  \textbf{Context} & We've reached the peak of human intelligence. \\
  \hline
  MTask  &  i think you're missing the point.\\
  S2S+LM &  i think that's a pretty good idea. \\
  StyleFusion & i don't know what's going on.\\
  DialoGPT    & It is going to get a little higher and higher every couple of months.\\
  \textsc{Style-DGPT}   & so what happens once we reach a constant state of quantum fluctuations?\\
  Human       & This is like pure consciousness thinking of consciousness, the rational endpoint of the enlightenment dialectic.\\
  \hline
  \hline
  \textbf{Context} & My rabbit does this every day.. not sure what he's thinking about.  \\
  \hline
  MTask  &  i'm not sure what is going on here.\\
  S2S+LM &  i think that's a cute cat.\\
  StyleFusion &  i'm guessing he's a male or female.\\
  DialoGPT    & He was thinking of lubing his ears.\\
  \textsc{Style-DGPT}   & This is a very good point and i also would like to discuss some other differences in the behavior of this species.\\
  Human       & some sort of inner monologue.\\
  \hline
  \end{tabular}
  \caption{The example outputs on the arXiv-style response generation task.}
  \label{tab:example_arXiv}
\end{table}

\begin{table}[th!]
  \centering
  \small
  \begin{tabular}{|p{0.08\textwidth}|p{0.34\textwidth}|} 
    \hline 
  \textbf{Context} & What is your favorite short story ? \\
  \hline
  MTask  &  the first time i saw this i thought it was a good idea to put a ring on it.\\
  S2S+LM &  the ending is a good thing and the most awesome thing i 've done\\
  StyleFusion & the ring of fire\\
  DialoGPT    & The Last Question by David Foster Wallace\\
  \textsc{Style-DGPT}   & 	The one about the day before the accident, about an old lady who used to live in a shack.\\
  Human       & The multitude of short stories that my wife writes\\
  \hline
  \hline
  \textbf{Context} & If your username was a movie, what would be the plot?   \\
  \hline
  MTask  &  the first harry potter movie.\\
  S2S+LM &  there's a lot of things about the movie.\\
  StyleFusion &  it's a trap!\\
  DialoGPT    & Probably The Prestige\\
\textsc{Style-DGPT} & a story of a mad scientist who goes in to investigate something and discovers he's trapped in a cave\\
  Human       & two lovers escape to the great northwest \\
  \hline
  \end{tabular}
  \caption{The example outputs on the Holmes-style response generation task.}
  \label{tab:example_Holmes}
\end{table}

\section{Conclusions}
We exploit the pre-trained language models on the stylized response generation task. To incorporate the style information from the non-parallel data into the generation model, we propose two learning objectives from word level and sentence level to steer the output distribution towards the desired style. Evaluation results on arXiv-style and Holmes-style response generation tasks indicate the effectiveness of the proposed approach.

\section*{Acknowledgment}
This work was supported in part by the National Natural Science Foundation of China (Grant Nos.U1636211, 61672081, 61370126), the Beijing Advanced Innovation Center for Imaging Technology (Grant No.BAICIT2016001), and the Fund of the State Key Laboratory of Software Development Environment (Grant No.SKLSDE2019ZX-17).

\bibliography{emnlp2020}
\bibliographystyle{acl_natbib}

\appendix

\end{document}